%% file: main.tex
\documentclass{article}

\PassOptionsToPackage{numbers, compress}{natbib}

\usepackage[final]{neurips_2022} 

\usepackage[utf8]{inputenc} 
\usepackage[T1]{fontenc}    
\usepackage{hyperref}       
\usepackage{url}            
\usepackage{booktabs}       
\usepackage{amsfonts}       
\usepackage{nicefrac}       
\usepackage{microtype}      
\usepackage{xcolor}         
\usepackage{graphicx}
\usepackage{lipsum}
\usepackage{chngpage}
\usepackage[para]{footmisc}

\input{macros.tex}

\title{A Modality-level Explainable Framework for Misinformation Checking in Social Networks}

\author{%
  V\'{i}tor Louren\c{c}o \\
  Universidade Federal Fluminense \\
  Dell Technologies\\
  Niter\'{o}i, Rio de Janeiro, Brazil \\
  \texttt{vitorlourenco@id.uff.br} \\
  \And
  Aline Paes \\
  Universidade Federal Fluminense \\
  Niter\'{o}i, Rio de Janeiro, Brazil \\
  \texttt{alinepaes@ic.uff.br} \\
}

\begin{document}

\maketitle

\begin{abstract}
    The widespread of false information is a rising concern worldwide with critical social impact, inspiring the emergence of fact-checking organizations to mitigate misinformation dissemination. However, human-driven verification leads to a time-consuming task and a bottleneck to have checked trustworthy information at the same pace they emerge. Since misinformation relates not only to the content itself but also to other social features, this paper addresses automatic misinformation checking in social networks from a multimodal perspective. Moreover, as simply naming a piece of news as incorrect may not convince the citizen and, even worse, strengthen confirmation bias, the proposal is a modality-level explainable-prone misinformation classifier framework. Our framework comprises a misinformation classifier assisted by explainable methods to generate modality-oriented explainable inferences. Preliminary findings show that the misinformation classifier does benefit from multimodal information encoding and the modality-oriented explainable mechanism increases both inferences' interpretability and completeness.
\end{abstract}

\section{Introduction\label{sec:intro}}
\input{sec/intro.tex}

\section{Methodology\label{sec:meth}}
\begin{figure}[t]
    \centering
    \includegraphics[width=\textwidth]{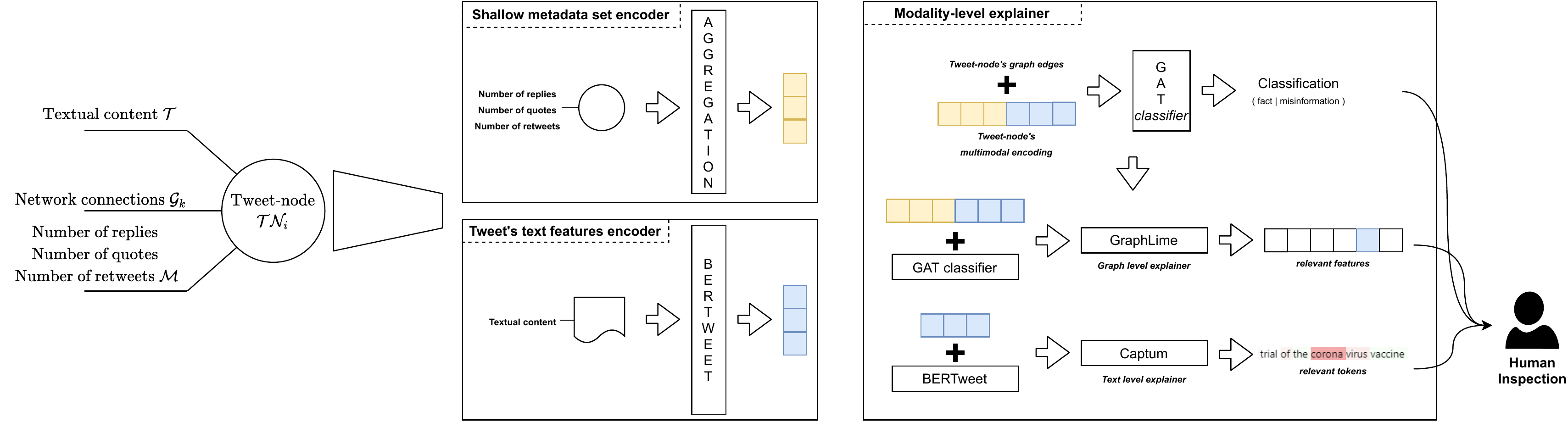}
    \caption{The proposed modality-level explainable framework. Red and green elements illustrate graph-based and text-based feature vectors, respectively.}
    \label{fig:framework}
\end{figure}

\input{sec/meth.tex}

\section{Experiments\label{sec:exp}}
\input{sec/exp.tex}

\vspace{-0.25cm}
\section{Conclusion\label{sec:conc}}
\vspace{-0.25cm}
\input{sec/conc.tex}

\begin{ack}
The authors would like to thank the Brazilian Research agencies FAPERJ and CNPq for the financial support.
\end{ack}


\newpage
\bibliographystyle{abbrvnat}
\small
\bibliography{refs}

\end{document}

%% file: macros.tex
\usepackage{soul}
\usepackage{xcolor}

\definecolor{cb-black}{RGB}{0,0,0}
\definecolor{cb-orange}{RGB}{230,159,0}
\definecolor{cb-skyblue}{RGB}{86,180,233}
\definecolor{cb-bluishgreen}{RGB}{0,158,155}
\definecolor{cb-yellow}{RGB}{240,228,66}
\definecolor{cb-blue}{RGB}{0,114,178}
\definecolor{cb-vermillion}{RGB}{213,94,0}
\definecolor{cb-reddishpurple}{RGB}{204,121,167}

\definecolor{darkgreen}{rgb}{0.0, 0.2, 0.13}

\newcommand{\eg}{\emph{e.g.}}
\newcommand{\ie}{\emph{i.e.}}



%% file: sec/intro.tex
The paper-free and web-based shift in the traditional press (\eg{}, CNN, Globo, The New York Times) has contributed to democratizing access to daily news. Another recent behavior shift of digital press is to leverage social media vehicles (\eg{}, Reddit, Twitter, Facebook, WhatsApp, TikTok) to spread information to time-critical news (\eg{}, life-threatening situations). For example, Pew Research Center's 2021 news consumption report~\cite{Walker2021} indicates that 48\% of Americans have social networks as a source of news, while Statista's News Consumption in Latin America Dossier~\cite{Statista2022} indicates that approximately two-thirds of people interviewed in Argentina, Brazil, Chile, and Mexico have social platforms as their news source.

With the advent of digital media, another observed trend is that everyone can share a piece of information. For example, the reports mentioned above do not separate the consumption of news coming from official presses or any other user. This way, the ease and agility with which information is disseminated on social networks come with their pitfalls, such as misleading information to totally false information. The widespread of false or misleading information (referred to as \textit{misinformation} in this paper) is a rising concern with critical social impact, as seen during the past elections and the COVID-19 pandemic~\cite{shu2017fake,Shu2018,Zhou2020,Shang2021,Shang2022}. Accordingly, the Statista's Dossier, Latin America is the most affected region with misinformation, varying from health and politics to products and services. 

In recent years, fact-checking organizations emerged (\eg{}, Lupa~\footnote{\url{https://lupa.uol.com.br/}}, Aos Fatos~\footnote{\url{https://www.aosfatos.org/}}, Chequeado~\footnote{\url{https://chequeado.com/}}) to mitigate misinformation dissemination. These organizations usually rely on multiple sources and documents -- \ie{}, multimodal, multilingual, and multi-topical data --, and human-driven verification to classify social media claims and news articles as trustworthy or misinformation, leading to a time-consuming task and, mostly, a bottleneck to overcome to have checked trustworthy information. To address the lack of scalability of traditional manual fact-checking, several work~\cite{Zhou2020} propose methods for automatic fact-checking social media content. For instance, \citet{Shang2021}~propose a multimodal approach for COVID-19-related misinformation detection on TikTok, \citet{shu2017fake,Shu2018} and \citet{Nielsen2022mumin} propose the \texttt{FakeNewsNet} and \texttt{MuMiN}, respectively, which both are resources that links claims with Twitter posts.

Although automatic fact-checking improves scalability, the aforementioned approaches lack a key aspect towards providing an ultimate tool for specialists or fact-checking agencies: explainability. Explanatory factors are decisive for users' confidence and to address legal concerns. Like so, recent work explore explainable methodologies to meet this requirement. As examples, \citet{Kai2019} and \citet{Kou2022} propose explainable frameworks for fake news detection, and \citet{Kou2020} and \citet{Shang2022} design solutions for multimodal explainable misinformation checking.

In this work, we address social networks' misinformation checking from a modality-level explainable perspective. Different from related work that do not tackle explainability from both \emph{interpretable} and \emph{complete} aspects~\cite{Gilpin2018}, our work combines two modality-oriented explainable methods to overcome the lack of interpretability from general multimodal solutions while keeping explanatory factors understandable (\ie{}, complete). In detail, the called \emph{modality-level explainable framework} aims to combine social networks' graph-based structure with their semantics and multimodal content targeting to classify social networks' misinformation posts while pointing out the explanatory factors that led to each classification. To do so, in our framework, we combine Graph Attention Network (GAT) classifier~\cite{velickovic2018graph}, a graph-based classier model, with GraphLime~\cite{Huang2020graphlime}, a graph-based explainable model, and Captum~\cite{kokhlikyan2020captum}, a text-based explainable tool, to the explained misinformation classification.
Our preliminary findings show that the combination of different data modalities improves overall classification and a qualitative inquiry over provided explanations shows that they do contribute to overall classification understanding.


%% file: sec/meth.tex
This section presents our modality-level explainable framework. Our goal with this framework is to provide a solution to detect misinformation from multimodal, multilingual, and multi-topical Twitter posts (also known as \textit{tweets}) and identify core features able to explain the classification, empowering human interpretation. We, first, state the misinformation tweet classification problem from a multimodal and explainable facet. Subsequently, we present our framework. 

\subsection{Problem statement}

Given a tweet-node $\mathcal{TN}_i = (\mathcal{T}, \mathcal{G}_k, \mathcal{M}, \mathcal{C})$, where $\mathcal{T}$ is the tweet itself (textual content); $\mathcal{G}_k$ defines the local $k$-hop network connections of the tweet, such as replies, quotes, and retweets (\ie{}, re-post of the same tweet); $\mathcal{M}$ is the tweet's shallow metadata set, which includes, as examples, the tweet's user owner, the number of likes, the location, among others; and $\mathcal{C}$ being the tweet's multimodal content, that might include images, videos, and audios. Our goal is to classify the tweet as \textit{misinformation} (\ie{}, $y_i = 0$) or \textit{fact} (\ie{}, $y_i = 1$). After, identify the most representative features within $\mathcal{TN}_i$ that led the classification.

\subsection{Modality-level explainable framework}

As illustrated in Figure~\ref{fig:framework}, our framework comprises two main steps: tweet-node encoding; and misinformation detection and classification explainability. The tweet-node encoding is divided twofold to encode the tweet-node's shallow metadata set $\mathcal{M}$, and the tweet's textual content $\mathcal{T}$. In this work, the shallow metadata features are a result of the aggregation of the number of replies, quotes, and retweets, while the text-based features are generated by encoding the textual content. After, both feature vectors are concatenated to form a unique multimodal tweet-node's vector representation.

The second main step is misinformation detection with its explainability. Aiming to accurately identify misinformation, we encode the tweet-node's graph connections $\mathcal{G}_k$. After, we trained the classifier to label tweets between fact or misinformation, coupled with the multimodal tweet-node's vector representation. Finally, to provide explanations over the associated label, we employ graph-based and text-based explainable methods. The graph-based method is able to identify the most important features within the tweet-node's multimodal representation, while the text-based method highlights textual aspects that lead to a better understating of the associated label.

%% file: sec/exp.tex
We conduct experiments on the MuMiN~\cite{Nielsen2022mumin} dataset, drawing preliminary insight into the proposed modality-level explainability of misinformation classification.

The MuMiN dataset is a public misinformation graph dataset that contains multimodal information from Twitter. Specifically, MuMiN associates multi-topical and multilingual tweets with fact-checked claims, and it also includes textual and visual content from tweets. We used the \texttt{MuMiN-small} version, which contains $2183$ claims and $7202506$ tweets. From the dataset, we filtered the English written tweets, four entities type (\emph{Claim}, \emph{Tweet}, \emph{Reply}, and \emph{User}), and six relations (\emph{Posted}, \emph{Mentions}, \emph{Retweeted}, \emph{Quote\_Of}, \emph{Reply\_To}, and \emph{Discusses}) to our analyses. Furthermore, we aim our efforts at the characterization of textual and relational features representation.

We developed our framework\footnote{\url{https://github.com/vitornl/MeLLL-MuMiN-explainable}\\}
using PyTorch~\cite{NEURIPS2019_9015}. As our misinformation classifier, we opt to use Graph Attention Networks~(GAT)~\cite{velickovic2018graph}, which is a widely used GNN model. The classifier was trained using a single Nvidia RTX2060 GPU, with a learning rate of $0.005$, $16$-dimensional hidden layer, and Adam optimizer~\cite{Kingma2015} for a total of $800$ epochs. The textual features were encoded using HuggingFace's\footnote{\url{https://huggingface.co/docs/transformers/model_doc/bertweet}} pretrained BERTweet model~\cite{Nguyen2020BERTweet}, followed by a linear mapping of its original 768-dimensional space vector to a 3-dimensional space vector to match the dimensions of the shallow metadata set representation. The dataset split used was the same proposed in the MuMiN dataset.

\begin{table}[t]
\centering
\caption{GAT's performance score in the misinformation classification task regarding each input feature. We report the F1-score average and standard deviation values from five independent executions.}
\label{tab:results}
\begin{tabular}{@{}lc@{}}
\toprule
\multicolumn{1}{c}{GAT's input tweet representation} & F1-score \\ \midrule
Graph-based features \textit{only}                 & 0.9225 \footnotesize{$\pm$ 0.0260}  \\
Text-based features \textit{only}                  & 0.8942 \footnotesize{$\pm$ 0.0097}  \\
\textbf{Multimodal features}                       & \textbf{0.9444 \footnotesize{$\pm$ 0.0052}}   \\ \bottomrule
\end{tabular}
\end{table}

\subsection{Preliminary Findings}

In the following, we present our preliminary findings. First, we quantitatively evaluate our misinformation classifier. To do so, we subject it to three different input scenarios: graph-based features \textit{only}, text-based features \textit{only}, and multimodal features, which is the concatenation of both graph-based and text-based features; and measure the obtained F1-score. In the second experiment, we perform a qualitative analysis of the identified explanatory features.

The results of the former experiment are shown in Table~\ref{tab:results} and are the average and standard deviation values from five independent executions. It demonstrates that the multimodal feature vector enables the GAT model to better generalize the domain and, as result, achieves better overall performance in misinformation detection, which confirms the results obtained by~\citet{Nielsen2022mumin}.

For the latter experiment, we qualitatively analyze the provided most important features and their capability of indeed correlating the tweet-node with its associated label. Our investigations are shown in Table~\ref{tab:investigation}, where the tweet-node text and shallow metadata set are displayed along with its GAT's classification, the GraphLime's~\cite{Huang2020graphlime} most representative features, and an interpretation of the explanatory factors. On the first entry, our GAT model classifies the tweet-node being as misinformation and GraphLime indicates that the most representative features are the number of retweets and the number of replies. After, analyzing both retweets and replies we identify that users that reply or retweet have other tweets classified as misinformation as well, suggesting a misinformation spread trend. The second entry was classified as being a fact while having the text embedding feature part being GraphLime's most representative feature. Like so, we further explore Captum's~\cite{kokhlikyan2020captum} textual factors that assist the label explanation. We observe that the ``protest against \#coronavirus'' and ``Let's see what happens to \#Germany" parts corroborate the correct classification.

\begin{table}[t]
\begin{adjustwidth}{-3.75cm}{0cm}
\centering
\caption{Report of two inferred cases with their classification, features, and explanatory factors identified by our framework.}
\vspace{-0.5cm}
\label{tab:investigation}
\resizebox{\paperwidth}{!}{%
\begin{tabular}{@{}llccl@{}}
\toprule
\multicolumn{1}{c}{Text} &
  Shallow metadata set &
  Classification &
  \begin{tabular}[c]{@{}c@{}}GraphLime\\ most representative features\end{tabular} &
  \multicolumn{1}{c}{Human interpretation} \\ \midrule
\begin{tabular}[c]{@{}l@{}}Great news! Carona virus vaccine ready. Able to cure patient within 3 hours\\ after injection. Hats off to US Scientists. Right now Trump announced that\\ Roche Medical Company will launch the vaccine next Sunday, and millions\\ of doses are ready from it !!! VIA: @wajih79273180 https://t.co/BZJCLtwuXq\end{tabular} &
  \begin{tabular}[c]{@{}l@{}}Number of retweets 26\\ Number of replies 42\\ Number of quotes 7\end{tabular} &
  Misinformation &
  \begin{tabular}[c]{@{}c@{}}Number of retweets\\ Number of replies\end{tabular} &
  \begin{tabular}[c]{@{}l@{}}Users who retweet tend\\ to spread misinformation\end{tabular} \\
  & & & & \\
\begin{tabular}[c]{@{}l@{}}17,000 anti-vaxxers, anti-science, far-right \&amp; neo-Nazi organizations\\ attend a protest against \#coronavirus restrictions in \#Berlin \&amp; defy\\ \#publichealth precautions. Let's see what happens to \#Germany's \#COVID19\\ case counts in the next 2-3 weeks. https://t.co/TD5xIoT5sV\end{tabular} &
  \begin{tabular}[c]{@{}l@{}}Number of retweets 11\\ Number of replies 9\\ Number of quotes 2\end{tabular} &
  Fact &
  Text &
  \begin{tabular}[c]{@{}c@{}}Word importance\\illustrated in Figure~\ref{fig:textexplainer}.\end{tabular}
  \\ \bottomrule
\end{tabular}
}
\end{adjustwidth}
\vspace{-0.3cm}
\end{table}

\begin{figure}
    \centering
    \includegraphics[width=\textwidth]{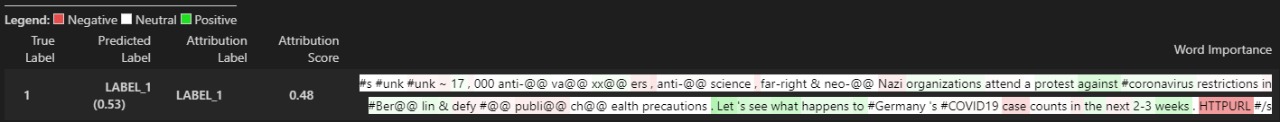}
    \caption{Word importance for text explainability of the second tweet from Table~\ref{tab:investigation}.}
    \label{fig:textexplainer}
    \vspace{-0.5cm}
\end{figure}

%% file: sec/conc.tex
In this paper, we devised the modality-level explainable framework, a solution to detect misinformation from Twitter posts and provide explanatory factors by identifying the most relevant features that lead to the classification. Our framework comprises two main steps: encoding tweet-node information and detecting possible misinformation posts while providing relevant features that guide the classification explanation. Our preliminary findings show that the proposed framework confirms previous misinformation classification results by achieving the best when exposed to the multimodal vector space. Besides, our qualitative analysis demonstrates how the modality-level does contribute to overall classification understanding. Namely, the graph-based explainer is able to identify the most relevant features within the text-node representation vector, amid the text-based explainer enhances the classification understating at the tweet level.

Our perception of future work is twofold. At first, our goal is to enhance and further explore the proposed modality-level frameworks by incorporating other modality-driven explainable methods. We hypothesize that leveraging more modality might enhance the misinformation classifier and, mainly, contribute to improving both explanations' interpretation and completeness. Secondly, we target to explore how each modality influences topic-specific subjects (\eg{}, COVID-19 pandemic, and elections).

%% file: main.bbl
\begin{thebibliography}{18}
\providecommand{\natexlab}[1]{#1}
\providecommand{\url}[1]{\texttt{#1}}
\expandafter\ifx\csname urlstyle\endcsname\relax
  \providecommand{\doi}[1]{doi: #1}\else
  \providecommand{\doi}{doi: \begingroup \urlstyle{rm}\Url}\fi

\bibitem[Gilpin et~al.(2018)Gilpin, Bau, Yuan, Bajwa, Specter, and
  Kagal]{Gilpin2018}
L.~H. Gilpin, D.~Bau, B.~Z. Yuan, A.~Bajwa, M.~Specter, and L.~Kagal.
\newblock Explaining explanations: An overview of interpretability of machine
  learning.
\newblock \emph{Proc. - 2018 IEEE 5th International Conference on Data Science
  and Advanced Analytics, DSAA 2018}, pages 80--89, 5 2018.

\bibitem[Huang et~al.(2020)Huang, Yamada, Tian, Singh, Yin, and
  Chang]{Huang2020graphlime}
Q.~Huang, M.~Yamada, Y.~Tian, D.~Singh, D.~Yin, and Y.~Chang.
\newblock Graphlime: Local interpretable model explanations for graph neural
  networks.
\newblock 1 2020.
\newblock ISSN 2331-8422.

\bibitem[Kingma and Ba(2015)]{Kingma2015}
D.~P. Kingma and J.~Ba.
\newblock Adam: {A} method for stochastic optimization.
\newblock In Y.~Bengio and Y.~LeCun, editors, \emph{3rd International
  Conference on Learning Representations, {ICLR} 2015}, 2015.

\bibitem[Kokhlikyan et~al.(2020)Kokhlikyan, Miglani, Martin, Wang, Alsallakh,
  Reynolds, Melnikov, Kliushkina, Araya, Yan, and
  Reblitz{-}Richardson]{kokhlikyan2020captum}
N.~Kokhlikyan, V.~Miglani, M.~Martin, E.~Wang, B.~Alsallakh, J.~Reynolds,
  A.~Melnikov, N.~Kliushkina, C.~Araya, S.~Yan, and O.~Reblitz{-}Richardson.
\newblock Captum: {A} unified and generic model interpretability library for
  pytorch.
\newblock \emph{CoRR}, abs/2009.07896, 2020.

\bibitem[Kou et~al.(2020)Kou, Yue~Zhang, Shang, and Wang]{Kou2020}
Z.~Kou, D.~Yue~Zhang, L.~Shang, and D.~Wang.
\newblock Exfaux: A weakly supervised approach to explainable fauxtography
  detection.
\newblock In \emph{2020 IEEE International Conference on Big Data (Big Data)},
  pages 631--636, 2020.

\bibitem[Kou et~al.(2022)Kou, Shang, Zhang, and Wang]{Kou2022}
Z.~Kou, L.~Shang, Y.~Zhang, and D.~Wang.
\newblock Hc-covid: A hierarchical crowdsource knowledge graph approach to
  explainable covid-19 misinformation detection.
\newblock \emph{Proc. ACM Hum.-Comput. Interact.}, 6\penalty0 (GROUP), jan
  2022.

\bibitem[Nguyen et~al.(2020)Nguyen, Vu, and Nguyenh]{Nguyen2020BERTweet}
D.~Q. Nguyen, T.~Vu, and A.~T. Nguyenh.
\newblock Bertweet: A pre-trained language model for english tweets.
\newblock pages 9--14. Association for Computational Linguistics (ACL), 11
  2020.

\bibitem[Nielsen and McConville(2022)]{Nielsen2022mumin}
D.~S. Nielsen and R.~McConville.
\newblock Mumin: A large-scale multilingual multimodal fact-checked
  misinformation social network dataset.
\newblock \emph{SIGIR 2022 - Proc. of the 45th International ACM SIGIR
  Conference on Research and Development in Information Retrieval}, pages
  3141--3153, 7 2022.

\bibitem[Paszke et~al.(2019)Paszke, Gross, Massa, Lerer, Bradbury, Chanan,
  Killeen, Lin, Gimelshein, Antiga, Desmaison, Kopf, Yang, DeVito, Raison,
  Tejani, Chilamkurthy, Steiner, Fang, Bai, and Chintala]{NEURIPS2019_9015}
A.~Paszke, S.~Gross, F.~Massa, A.~Lerer, J.~Bradbury, G.~Chanan, T.~Killeen,
  Z.~Lin, N.~Gimelshein, L.~Antiga, A.~Desmaison, A.~Kopf, E.~Yang, Z.~DeVito,
  M.~Raison, A.~Tejani, S.~Chilamkurthy, B.~Steiner, L.~Fang, J.~Bai, and
  S.~Chintala.
\newblock Pytorch: An imperative style, high-performance deep learning library.
\newblock In H.~Wallach, H.~Larochelle, A.~Beygelzimer, F.~d\textquotesingle
  Alch\'{e}-Buc, E.~Fox, and R.~Garnett, editors, \emph{Advances in Neural
  Information Processing Systems 32}, pages 8024--8035. Curran Associates,
  Inc., 2019.

\bibitem[Shang et~al.(2021)Shang, Kou, Zhang, and Wang]{Shang2021}
L.~Shang, Z.~Kou, Y.~Zhang, and D.~Wang.
\newblock A multimodal misinformation detector for covid-19 short videos on
  tiktok.
\newblock In \emph{2021 IEEE International Conference on Big Data (Big Data)},
  pages 899--908, 2021.

\bibitem[Shang et~al.(2022)Shang, Kou, Zhang, and Wang]{Shang2022}
L.~Shang, Z.~Kou, Y.~Zhang, and D.~Wang.
\newblock A duo-generative approach to explainable multimodal covid-19
  misinformation detection.
\newblock \emph{Proc. of the ACM Web Conference - WWW 2022}, pages 3623--3631,
  4 2022.

\bibitem[Shu et~al.(2017)Shu, Sliva, Wang, Tang, and Liu]{shu2017fake}
K.~Shu, A.~Sliva, S.~Wang, J.~Tang, and H.~Liu.
\newblock Fake news detection on social media: A data mining perspective.
\newblock \emph{ACM SIGKDD Explorations Newsletter}, 19\penalty0 (1):\penalty0
  22--36, 2017.

\bibitem[Shu et~al.(2018)Shu, Mahudeswaran, Wang, Lee, and Liu]{Shu2018}
K.~Shu, D.~Mahudeswaran, S.~Wang, D.~Lee, and H.~Liu.
\newblock Fakenewsnet: {A} data repository with news content, social context
  and dynamic information for studying fake news on social media.
\newblock \emph{CoRR}, abs/1809.01286, 2018.

\bibitem[Shu et~al.(2019)Shu, Cui, Wang, Lee, and Liu]{Kai2019}
K.~Shu, L.~Cui, S.~Wang, D.~Lee, and H.~Liu.
\newblock Defend: Explainable fake news detection.
\newblock In \emph{Proc. of the 25th ACM SIGKDD International Conference on
  Knowledge Discovery \& Data Mining}, KDD '19, page 395–405. ACM, 2019.

\bibitem[Statista(2022)]{Statista2022}
Statista.
\newblock News consumption in latin america | statista, 2022.
\newblock URL \url{https://www.statista.com/topics/8083/news-in-latin-america}.

\bibitem[Veli{\v{c}}kovi{\'{c}} et~al.(2018)Veli{\v{c}}kovi{\'{c}}, Cucurull,
  Casanova, Romero, Li{\`{o}}, and Bengio]{velickovic2018graph}
P.~Veli{\v{c}}kovi{\'{c}}, G.~Cucurull, A.~Casanova, A.~Romero, P.~Li{\`{o}},
  and Y.~Bengio.
\newblock {Graph Attention Networks}.
\newblock \emph{International Conference on Learning Representations}, 2018.

\bibitem[Walker and Matsa(2021)]{Walker2021}
M.~Walker and K.~E. Matsa.
\newblock News consumption across social media in 2021 | pew research center,
  2021.
\newblock URL
  \url{https://www.pewresearch.org/journalism/2021/09/20/news-consumption-across-social-media-in-2021/}.

\bibitem[Zhou and Zafarani(2020)]{Zhou2020}
X.~Zhou and R.~Zafarani.
\newblock A survey of fake news: Fundamental theories, detection methods, and
  opportunities.
\newblock \emph{ACM Comput. Surv.}, 53\penalty0 (5), sep 2020.

\end{thebibliography}
